%% file: neurips_2021.tex
\pdfoutput=1

\documentclass{article}

\PassOptionsToPackage{numbers, compress}{natbib}

    \usepackage[preprint]{neurips_2021}

\usepackage[utf8]{inputenc} 
\usepackage[T1]{fontenc}    
\usepackage{hyperref}       
\usepackage{url}            
\usepackage{booktabs}       
\usepackage{multirow}
\usepackage{amsfonts}       
\usepackage{nicefrac}       
\usepackage{microtype}      
\usepackage{xcolor}         
\usepackage{natbib}

\usepackage{amsmath}

\usepackage{graphicx}

\usepackage{wrapfig}

\usepackage{algorithm}
\usepackage{algorithmicx}  
\usepackage{algpseudocode}  
\usepackage{amsmath}  

\usepackage{amssymb}
\makeatletter

\newcommand{\Rmnum}[1]{\expandafter\@slowromancap\romannumeral #1@}
\makeatother

\title{Backdoor Attacks on Federated Lottery Ticket Learning}
\title{Backdoor Attacks on Federated Learning with Lottery Ticket Hypothesis}

\author{
    Zeyuan Yin, Ye Yuan, Panfeng Guo, Pan Zhou\\
    Huazhong University of Science and Technology \\
    \texttt{\{zeyuanyin,maxwell\_yuan,panfengguo,panzhou\}@hust.edu.cn} \\
}

\begin{document}

\maketitle

\input{main_0914}

\clearpage
\bibliographystyle{unsrtnat}
\bibliography{references}

\clearpage
\input{appendix}

\end{document}

%% file: main_0914.tex
\begin{abstract}
Edge devices in federated learning usually have much more limited computation and communication resources compared to servers in a data center. 
Recently, advanced model compression methods, like the Lottery Ticket Hypothesis, have already been implemented on federated learning to reduce the model size and communication cost.
However, Backdoor Attack can compromise its implementation in the federated learning scenario. The malicious edge device trains the client model with poisoned private data and uploads parameters to the center, embedding a backdoor to the global shared model after unwitting aggregative optimization. 
During the inference phase, the model with backdoors classifies samples with a certain trigger as one target category, while shows a slight decrease in inference accuracy to clean samples. 
In this work, we empirically demonstrate that Lottery Ticket models are equally vulnerable to backdoor attacks as the original dense models, and backdoor attacks can influence the structure of extracted tickets. 
Based on tickets' similarities between each other, we provide a feasible defense for federated learning against backdoor attacks on various datasets. 
Codes are available at \url{https://github.com/zeyuanyin/LTH-Backdoor}.

\end{abstract}

\vspace{-0.5em}
\section{Introduction}
\label{introduction}
\vspace{-0.3em}
Federated learning (FL) has been proposed to address the privacy problems without direct access to sensitive training data, especially for privacy-sensitive tasks \citep{yang2019federated,kairouz2019advances}. 
However, edge devices usually have much more limited computation and communication resources compared to the traditional data center, which means that the client model should be as light as possible. 
Lots of model compression methods are proposed to solve this problem \citep{han2015learning,guo2016dynamic,molchanov2016pruning,liu2018rethinking,liu2017learning}. 
One of them, the Lottery Ticket Hypothesis (LTH) \citep{frankle2019lottery} shows that there exist subnetworks (lottery tickets) that can reach test accuracy comparable to the original network with fewer computational resources usage.
\citet{li2020lotteryfl} proposed LotteryFL which applied LTH in FL scenario to alleviate the scarcity of computational resources on edge devices and reduce the communication cost. 
Nevertheless, various security issues from the real world still need to be tackled \citep{goodfellow2014explaining,bagdasaryan2020backdoor,biggio2018wild}. 
Backdoor attacks, which are launched during the training phase, have been recently studied on FL \citep{bagdasaryan2020backdoor,xie2019dba}.
If the attackers hack into some clients and maliciously modify their datasets, the models uploaded from the infected clients can compromise the global shared model, then the vulnerable FL system may have a high risk of data leakage.

In this paper, 
we empirically demonstrate that Lottery Ticket model is equally vulnerable to the backdoor attack as the original dense model.
We provide detailed experiments and analysis about the phenomenon based on the theory of \citet{liu2018fine} that backdoor attacks embed backdoors in neurons or connections of neurons. 
Neurons with decisive roles on the high test accuracy are defined as key neurons. 
We identify the existence of key neurons in both normal tickets and backdoor tickets by contrasting the similarity between benign pruned models and backdoor pruned models, which helps to explain the equal vulnerability. 
Furthermore, we design a detection method to detect the backdoor tickets and apply this method to the aggregative optimization process of FL to enhanced robustness of global shared models in real-world deployments. 

Our contributions are summarized as follows:
\vspace{-.5em}
\begin{enumerate}
\setlength{\itemsep}{-0.5pt } 
\item We conduct various backdoor attacks on the essence of LotteryFL, Lottery Ticket Hypothesis, and demonstrate that the lottery ticket is vulnerable, which arguably shows that backdoor attacks affect lottery tickets' structures and bring potential security risks to practical application.

\item We discuss the principle of backdoor embedding and the intrinsic mechanism of backdoor learning in detail, showing that neurons, retained in benign tickets, also have a high probability of being retained in backdoor tickets.

\item We propose a feasible method for LotteryFL to detect backdoor attacks on various datasets, which can develop an efficient and robust (against backdoor attacks) FL system.

\end{enumerate}

    \vspace{-1.3em}
\section{Related Works}
\label{related_works}
    \vspace{-0.7em}
\textbf{Lottery Ticket Hypothesis.}
\citet{frankle2019lottery} first demonstrate Lottery Ticket Hypothesis that there exists a small subnetwork, called the \textit{Lottery Ticket}, which can make the training process more efficient and lead to a comparable accuracy to the dense network. 
\citet{liu2018rethinking} thinks that it is not necessary to inherit weight from a large model and inheriting might trap the pruned model into a bad local minimum. Based on the hypothesis, 
\citet{you2020drawing} improve the searching for lottery tickets and propose \textit{Early-Bird Tickets}, a kind of lottery tickets which can be drawn at very early iterations.

\textbf{Federated Learning.}
Federated learning enables model training from local data collected by edge/mobile devices by uploading, downloading and aggregating model parameters, while preserving data privacy \citep{yang2019federated,kairouz2019advances}. 
In terms of accelerating FL model training, PruneFL \citep{jiang2019model} had been proposed to reduce both communication and computation overhead and minimize the overall training time with parameter pruning. 
LotteryFL \citep{li2020lotteryfl} is a personalized and communication-efficient federated learning framework via exploiting the Lottery Ticket hypothesis.

\textbf{Backdoor Attacks.}
Backdoor attack \citep{kairouz2019advances,chen2017targeted}, an artful kind of poison attack \citep{mei2015using,jagielski2018manipulating}, is to embed backdoors into a model by training with poisoning data.
The backdoor can hide and does not affect the performance on no-trigger data, which makes users hard to judge the existence of the backdoor only by the accuracy of the validation dataset. If the input data is with a particular trigger, the backdoor will be activated and contribute to misleading the classification result to the attack target, which will limit the using range of the model due to potential security risk.
Threats from backdoor attacks are widespread in the FL field \citep{bagdasaryan2020backdoor,xie2019dba}. 
In many practical scenarios, a minority attacked clients will embed backdoors on the FL model and disturb the FL application, which shows that FL is vulnerable to the backdoor attack.


    \vspace{-1em}
\section{Approach}
\label{approach}
    \vspace{-0.7em}
\subsection{Backdoor Attacks on LotteryFL}
\vspace{-0.7em}
\textbf{Backdoor Attacks.}
Inspired by backdoor attack variants \citep{wang2019neural}, we use two general attack methods to evaluate the LotteryFL and our proposed defense method. 
\textit{BadNet Attack} \citep{gu2017badnets} is a classic attack with a white square trigger $k$ injected in the bottom right corner of each poisoning image. 
Besides, we design another attack form called \emph{Random Trigger Attack} by replacing the white square trigger with a random square trigger $k_{rand}$.
The evaluation of backdoor attacks are based on two metrics, Clean Data Accuracy (CDA) and Attack Success Rate (ASR)\citep{gao2020backdoor}, specifically defined in appendix \ref{appendix:backdoor}.

\textbf{Attack on LTH.}
Backdoor attacks on LotteryFL are essentially attacks on LTH, therefore we empirically evaluate the impacts of backdoor attacks on lottery tickets (LT). 
We study the robustness of Early-Bird Tickets \citep{you2020drawing}, a variant of LTs that can be drawn at very early iterations. 
We explore the structural difference between LTs with / without backdoor attacks, 
and analyze the the causes of high classification performance and successful backdoor attack on the backdoor LTs.



\vspace{-0.7em}
\subsection{The Existence of Key Neurons}
\vspace{-0.7em}
We suppose most key neurons with decisive roles on the CDA will be preserved whether or not backdoors embedded into DNNs. 
This is corresponding to the Neural Architecture Search \citep{elsken2019neural} which is to find the optimal network structure in dense models. 
We conduct the experiments to estimate extent of that backdoor attacks affect the lottery tickets' architecture. 
In experiments, all random seeds are fixed to ensure that 1) the setting and results of each repeated experiment are the same, so the comparison of two benign tickets obtained by repeated experiments shows 100\% similarity; 2) the only change in the comparison experiment is whether there is backdoor data in the dataset $\mathcal{D}_{train}$.

\vspace{-.5em}
\input{algorithm}

\vspace{-0.84em}
\subsection{Defense on LotteryFL}
\vspace{-0.7em}
According to the larger difference on similarity between benign tickets and backdoor tickets than difference among benign tickets, 
we use anomaly detection \textbf{Detect()} \citep{chandola2009anomaly} based on the mask similarity to detect whether a client ticket is a backdoor ticket. 
$\mathcal{L} \in R^k$ denotes a boolean vector, where $\mathcal{L}[i] = true$ represents the detected backdoor ticket $\theta_i$ and $\mathcal{L}[i] = false$ represents the benign ticket $\theta_i$.
We use \textbf{FineTune()} to train the detected backdoor tickets $\theta_i[\mathcal{L}]$ for some epochs to mitigate the backdoors on server-own validation data, based on the effective fine-tuning methods in \citep{liu2018fine}.
For specific LotteryFL functions shown on Algorithm~\ref{algorithm}, \textbf{Prune()} outputs the mask $m_i$ by pruning $\theta_i$ at the pruning rate $r_p$;
\textbf{ClientUpdate()} distributes and receives weights from clients; 
\textbf{AggregateLTNs()} is an aggregative optimization process, like FedAvg \citep{mcmahan2017communication}.

\vspace{-1.2em}
\section{Experiments}
\label{exp}

\vspace{-0.9em}
\subsection{Explore the performance of backdoor LT}
\vspace{-0.7em}


\begin{table}[b]
    \vspace{-1em}
\centering
\caption{CDA and ASR of Lottery Tickets under different pruning rate.}
\label{preformance_of_ALT}
\resizebox{\textwidth}{!}{%
\begin{tabular}{clcccccccc}
\hline
\multicolumn{2}{c}{\multirow{2}{*}{\textbf{Setting}}} & \multicolumn{4}{c}{\textbf{Clean Data Accuracy (\%)}} & \multicolumn{4}{c}{\textbf{Attack Success Rate (\%)}} \\ \cline{3-10} 
\multicolumn{2}{c}{} & p=0 & p=0.3 & p=0.5 & p=0.7 & p=0 & p=0.3 & p=0.5 & p=0.7 \\ \hline
\multirow{3}{*}{\begin{tabular}[c]{@{}c@{}}VGG16\\ CIFAR-10\end{tabular}} & LT (benign) & 93.28 & 93.25 & 93.13 & \multicolumn{1}{c|}{92.60} & - & - & - & - \\
 & Backdoor (\Rmnum{1}) & 92.55 & 92.42 & 92.38 & \multicolumn{1}{c|}{92.34} & 95.59 & 96.36 & 96.60 & 96.69 \\
 & Backdoor (\Rmnum{2}) & 93.02 & 92.90 & 92.62 & \multicolumn{1}{c|}{91.98} & 100 & 100 & 100 & 100 \\ \hline
\multirow{3}{*}{\begin{tabular}[c]{@{}c@{}}VGG16\\ CIFAR-100\end{tabular}} & LT (benign) & 72.04 & 71.34 & 70.53 & \multicolumn{1}{c|}{69.91} & - & - & - & - \\
 & Backdoor (\Rmnum{1}) & 69.97 & 69.12 & 67.73 &  \multicolumn{1}{c|}{63.00} & 87.50 & 89.13 & 90.29 & 90.59  \\
 & Backdoor (\Rmnum{2}) & 69.48 & 68.08 & 67.93 &  \multicolumn{1}{c|}{64.62} & 99.78 & 99.92 & 100 & 100 \\ \hline
\multirow{3}{*}{\begin{tabular}[c]{@{}c@{}}RESNET18\\ CIFAR-10\end{tabular}} & LT (benign) & 92.71 & 92.93 & 92.93 & \multicolumn{1}{c|}{93.05} & - & - & - & - \\
 & Backdoor (\Rmnum{1}) & 90.66 & 91.02 & 91.34 &  \multicolumn{1}{c|}{90.06} & 97.28 & 97.33 & 97.35 &  97.22  \\
 & Backdoor (\Rmnum{2}) & 91.13 & 89.89 & 90.51 &  \multicolumn{1}{c|}{90.18} & 100 & 100 & 100 & 100 \\ \hline
\multirow{3}{*}{\begin{tabular}[c]{@{}c@{}}RESNET18\\ CIFAR-100\end{tabular}} & LT (benign) & 71.59 & 71.45 & 70.94 & \multicolumn{1}{c|}{70.04} & - & - & - & - \\
 & Backdoor (\Rmnum{1}) & 69.46 & 68.60 & 67.11 & \multicolumn{1}{c|}{65.04} & 89.03 & 87.59 & 89.73 & 89.67  \\
 & Backdoor (\Rmnum{2}) & 70.02 & 69.52 & 67.58 &  \multicolumn{1}{c|}{66.58} & 99.80 & 99.65 & 99.24 & 99.86  \\ \hline
\end{tabular}%
}
    \vspace{-2em}
\end{table}

In this section, we evaluate the performance of backdoor-embedded LT via CDA and ASR metrics. 
We retrain backdoor LTs with benign datasets to measure their CDA. Meanwhile, we retrain backdoor LTs with backdoor datasets to measure their ASR. 
In the \tablename~\ref{preformance_of_ALT}, \textit{Backdoor (\Rmnum{1})} and \textit{Backdoor (\Rmnum{2})} represent the \emph{White Trigger attacks} and \emph{Random Trigger attacks} separately.

\vspace{-0.2em}

We draw two main conclusions from the experiment data in \tablename~\ref{preformance_of_ALT}.
For CDA, comparing benign LT with backdoor LT at the same pruning rate $p$, we find that the latter is not significantly different from the former. For example, the CDA of Backdoor(\Rmnum{1}) is only 0.83\% lower than the benign LT. This demonstrates the highly concealed nature of the embedded backdoor. 
For ASR, comparing dense network $(p=0)$ with pruned network $(p>0)$ under the same attack, we find that there is high ASR in both dense and pruned networks, which indicates that the lottery ticket networks are equally vulnerable to backdoor attacks as the original dense networks.


\vspace{-1.2em}
\subsection{Identify Key Neurons}
\begin{figure}[t]
    \vspace{-2em}
    \centering
    \includegraphics[width=1\textwidth]{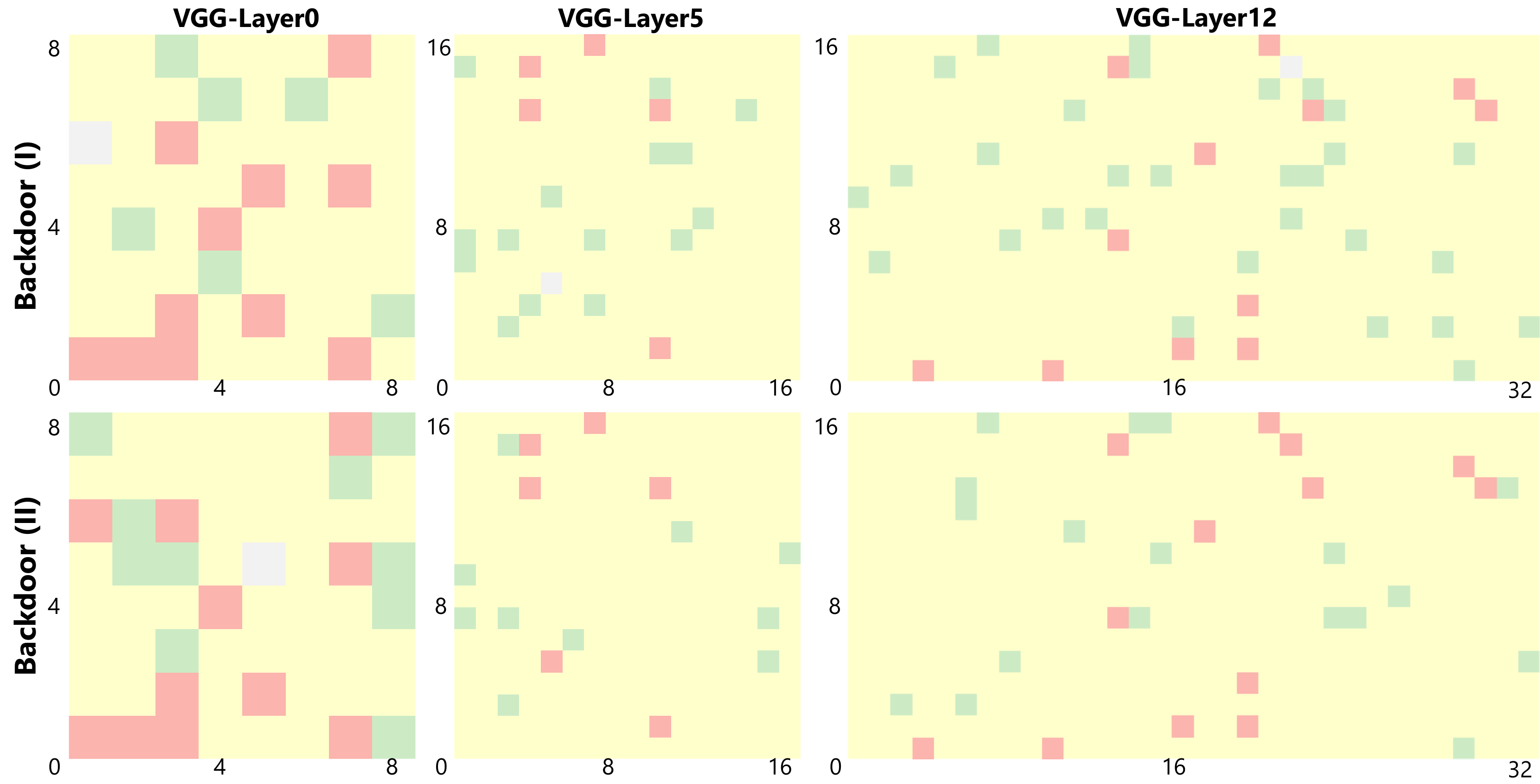}
        \vspace{-2em}
    \caption{Visualization of the  distribution heat map of retained neurons in $[0, 5, 12]^{th}$ layers between the normal model $f(m \odot \theta)$ and backdoor model $f^b(m\odot \theta)$ on CIFAR-10 where $p$ = 0.3.}
    \label{visualize}
    \vspace{-1.7em}
\end{figure}

\vspace{-0.8em}
To illustrate the causes of the high CDA on backdoor tickets, \figurename~\ref{visualize} plots the distribution of pruned neurons between the clean model $f(\theta)$ (trained by benign datasets $\mathcal{D}$), and backdoor model $f^b(\theta)$ (trained by backdoor datasets $\mathcal{D}^{b}$). 
Green squares and red squares represent the neurons retained in $f(m\odot \theta)$ and $f^b(m\odot \theta)$, respectively, where $m$ is a mask. Yellow squares stand for neurons that are both retained, and pale squares stand for neither retained.
\figurename~\ref{visualize} shows yellow squares occupy most of the position.
We can conclude that neurons that are retained in $f(m\odot \theta)$ also have a high probability of being retained in $f^b(m\odot \theta)$, thus the embedding of backdoors does not significantly compromise the model's CDA. 
In the appendix \ref{appendix:key_neurons}, we carry out further experiments to demonstrate how these both-retained neurons have decisive roles in the inference phase by precise figures.

\begin{wraptable}{r}{0.5\textwidth}
\centering
    \vspace{-0.5em}
    \caption{Contrasting the similarity of different tickets' structures with the same initialization seed.}
    \label{similarity_of_LT}

\resizebox{0.5\textwidth}{!}{

    \begin{tabular}{cllll}
    \hline
    \multirow{2}{*}{\textbf{Setting}} & \multicolumn{1}{c}{\multirow{2}{*}{\textbf{Compare}}} & \multicolumn{3}{c}{\textbf{Similarity}} \\ \cline{3-5} 
     & \multicolumn{1}{c}{} & p=30\% & p=50\% & p=70\% \\ \hline
    \multirow{3}{*}{\begin{tabular}[c]{@{}c@{}}VGG16\\ CIFAR-10\end{tabular}} & Benign - Backdoor (\Rmnum{1}) & 67.09 & 71.12 & 77.37 \\
     & Benign - Backdoor (\Rmnum{2}) & 67.52 & 72.40 & 77.98 \\ \cline{2-5} 
     & \textbf{Avg. Decrease} & \textbf{32.70} & \textbf{28.24} & \textbf{22.33} \\ \hline
    \multirow{3}{*}{\begin{tabular}[c]{@{}c@{}}VGG16\\ CIFAR-100\end{tabular}} & Benign - Backdoor (\Rmnum{1}) & 66.62 & 75.80 & 74.29 \\
     & Benign - Backdoor (\Rmnum{2}) & 66.52 & 76.52 & 77.32 \\ \cline{2-5} 
     & \textbf{Avg. Decrease} & \textbf{33.43} & \textbf{23.84} & \textbf{24.20} \\ \hline
    \multirow{3}{*}{\begin{tabular}[c]{@{}c@{}}RESNET18\\ CIFAR-10\end{tabular}} & Benign - Backdoor (\Rmnum{1}) & 67.42 & 58.75 & 58.25 \\
     & Benign - Backdoor (\Rmnum{2}) & 66.54 & 59.96 & 62.13 \\ \cline{2-5} 
     & \textbf{Avg. Decrease} & \textbf{33.02} & \textbf{40.64} & \textbf{39.81} \\ \hline
    \multirow{3}{*}{\begin{tabular}[c]{@{}c@{}}RESNET18\\ CIFAR-100\end{tabular}} & Benign - Backdoor (\Rmnum{1}) & 79.67 & 78.04 & 75.46 \\
     & Benign - Backdoor (\Rmnum{2}) & 79.17 & 78.20 & 75.08 \\ \cline{2-5} 
     & \textbf{Avg. Decrease} & \textbf{20.58} & \textbf{21.88} & \textbf{24.73} \\ \hline
    \end{tabular}

}

\vspace{-1em}
\end{wraptable}

\tablename~\ref{similarity_of_LT} shows the similarity between benign tickets and backdoor tickets. 
We observe that Benign - Backdoor Similarity is obviously lower than the Benign - Benign Similarity (100\%). Taking VGG16 with dataset CIFAR-10 under the backdoor attack (\Rmnum{1}) for example, the similarity value in the pruning 30\%, 50\%, 70\% decreases by 32.70\%, 28.24\%, 22.33\% on average respectively. 
We can conclude that the backdoor attack affect the structure of extracted lottery tickets during the searching tickets stage.

\vspace{-0.5em}
\begin{wrapfigure}{r}{0.45\textwidth}
\vspace{-1.5em}
\includegraphics[width=0.45\textwidth]{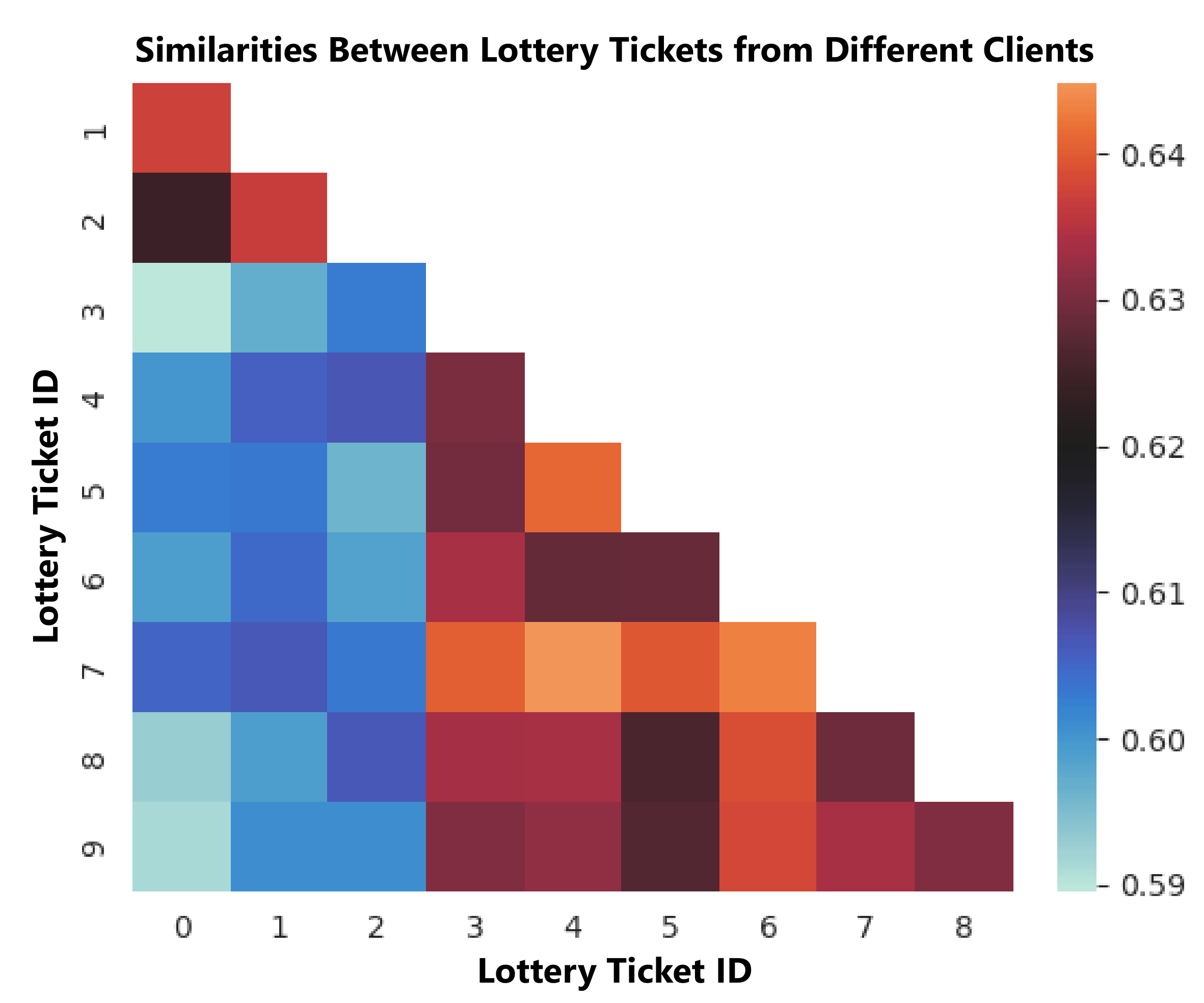}
\vspace{-2em}
\caption{The heat-map shows the similarity among different lottery tickets drawn from benign and malicious clients.}
\label{similar_hm}
\vspace{-1em}
\end{wrapfigure}

\subsection{Detect backdoor tickets on LotteryFL}
\label{Detect_LTFL}
\vspace{-0.5em}
We carry out experiments to verify the effectiveness of \textbf{Detect()} in Algorithm~\ref{algorithm}. We equally divide the CIFAR-10 dataset into 10 subsets, and mix a few subsets with backdoor data to form backdoor subsets. We train 10 client models on these 10 subsets respectively, and compare the similarity between 10 lottery tickets. 
The similarity score is calculated via the Hamming distance between any two tickets' masks. 
\figurename~\ref{similar_hm} illustrates the similarity scores between $[0,1,2]^{th}$ tickets and $[3-9]^{th}$ tickets are visibly lower than others, which shows that $[0,1,2]^{th}$ tickets are probably backdoor tickets. Then, these detected backdoor tickets will be processed in \textbf{FineTune()} to mitigate backdoors \citep{liu2018fine}. 
Even if \textbf{Detect()} may occasionally treats benign LT as backdoor LT by mistake, there is no harm other than consuming a little more computational resources, given that \textbf{FineTune()} can also be used to restore performance of the normal LT \citep{han2015learning}.


\section{Conclusion and Future Work}
\label{conclusion}
\vspace{-0.8em}
In this paper, we are the first to demonstrate the LotteryFL is as vulnerable to backdoor attacks as the origin FL and analyze in detail that the ticket's structure can be affected by the attack while most decisive neurons still retain under the attack. We further provide an integral defense algorithm on LotteryFL. 
There are promising research directions in the future: 1) robustness of LotteryFL should be evaluated by more advanced backdoor attacks; 2) more effective defense methods need to be proposed to detect backdoors in federated learning with label-deficient non-iid datasets.



%% file: algorithm.tex
\begin{wrapfigure}{r}{0.5\textwidth}
\vspace{-2.5em}
\begin{minipage}{\linewidth}

\begin{algorithm}[H]
    \caption{Backdoor Defense on LotteryFL}
    \label{algorithm}
    \begin{algorithmic}
    \Require 
    \State initialize the global model $\theta_g$
        \State $S_t \leftarrow \{C_1,\dots,C_k\}$ 
        \For{each round $t = 1, 2, \dots$}
            \For{each client $C_i \in S_t$ \textbf{in parallel}}
                \State $m_i^{t} \leftarrow$ \textbf{Prune}($\theta_{i}^t$, $r_p$)
                \State $\theta_{i}^t \leftarrow \theta_g^t\odot m_i^t$
                \State $\theta_{i}^{t+1} \leftarrow \textbf{ClientUpdate}(C_i, \theta_{i}^t)$ 
            \EndFor
            \State $\mathcal{L}^{t+1} \leftarrow \textbf{Detect}(\theta^{t+1})$
            \State $\theta^{t+1}[\mathcal{L}^{t+1}] \leftarrow \textbf{FineTune}(\theta^{t+1}[\mathcal{L}^{t+1}])$
            \State $\theta_g^{t+1} \leftarrow \textbf{AggregateLTNs}(\theta^{t+1})$
        \EndFor

    \end{algorithmic}
\end{algorithm}

\end{minipage}
\vspace{-1em}
\end{wrapfigure}

%% file: appendix.tex
\appendix

\section{Hyperparameter}

This Appendix includes training details of experiments in our paper (e.g., datasets, training models, hyperparameters).

\subsection{VGG16/ResNet18 on CIFAR-10/100}
The VGG16\citep{simonyan2014very} consists of 13 convolutional layers and 3 full connection layers.
The ResNet18\citep{he2016deep} is a 20 layer convolutional network with residual connections designed for CIFAR-10/100.

The CIFAR-10 dataset consists of 60000 32$\times$32 color images that are labeled with one of 10 classes. There are 6000 images per class with 5000 training and 1000 testing images per class. The CIFAR-100 dataset consists of 60000 32$\times$32 color images that are labeled with one of 100 classes. There are 500 training images and 100 testing images per class. The 100 classes in the CIFAR-100 are grouped into 20 superclasses. Each image comes with a "fine" label (the class to which it belongs) and a "coarse" label (the superclass to which it belongs).\footnote{\url{https://www.cs.toronto.edu/~kriz/cifar.html}}

We follow the training experimental setting of \citet{you2020drawing,frankle2019lottery}:
\begin{itemize}
\item We use the original splits of CIFAR-10/100 where 50000 training images are regarded as a train dataset $\mathcal{D}_{train}$, 10000 testing images are regarded as a test dataset $\mathcal{D}_{test}$. $\mathcal{D}_{val}$.
\item We use a batch size of 256.
\item We use batch normalization.
\item We use the optimization technique SGD with with momentum of 0.9.
\item We use the channel prune method and use the pruning rate $p$ from [0.3, 0.5, 0.7].
\item We use two learning rate schedules [0$_{LR\rightarrow0.1}$, 80$_{LR\rightarrow0.01}$, 120$_{LR\rightarrow0.001}$] and [0$_{LR\rightarrow0.1}$,70$_{LR\rightarrow0.01}$, 130$_{LR\rightarrow0.001}$].
\item We use the total iteration times of 160.

\end{itemize}

\subsection{Backdoor Attacks}
\label{appendix:backdoor}

The scenario of a successful attack is that the percentage of backdoor instances classified as the target label is high and the accuracy on the pristine test data of the poisoned model should be similar to the test accuracy of the pristine model \citep{chen2017targeted}. This means the backdoor model behaves normally for inputs containing no trigger, making it impossible to distinguish the backdoor model from the clean model by solely checking the test accuracy with the test samples.

\begin{figure}[H]
  \centering
  \includegraphics[width=1\textwidth]{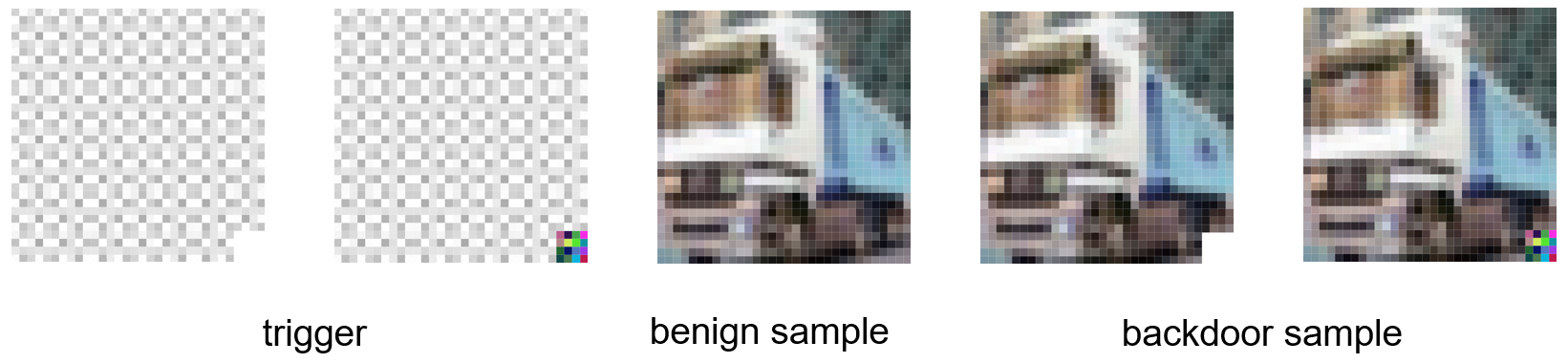}
     \vspace{-1.em}
  \caption{Examples of poisoning samples generated by BadNet Attack and Complex Trigger Attack. Left: two kinds of triggers $k$. Middle: benign sample $x$. Right: two kinds of backdoor sample $x^b$. Note that the pale grids are about $(i,j)\notin R$ and pixel value is 0 (invalid).}
  \label{backdoor_sample}
\end{figure}

Taking CIFAR-10 for an example, a dataset consisting of 32$\times$32 color images with the trigger $k$ embedded at the lower-left 4$\times$4 pixels position of the picture. The outer pixels' value is set to 0, which means the pixels are invalid. After obtaining the trigger $k$, the process of generating backdoor data $x^b$ is expressed as
$x^b_{i,j}=replace(x,k)
=\begin{cases}
x_{i,j}& \text{$k_{i,j}=0$}\\
k_{i,j}& \text{$k_{i,j}\neq0$}
\end{cases}$
where $replace(x,k)$ is used to replace pixel value of $x$ with the trigger $k$ when trigger pixel $k_{i,j}$ is valid. After generating a set of backdoor samples, we blend them with the entire benign train dataset $\mathcal{D}_{train}$ to form a malicious attack train dataset $\mathcal{D}^{b}_{train}$.
\figurename~\ref{backdoor_sample} shows the whole process of generating backdoor samples.

\vspace{-1em}
\textbf{Evaluation metrics.} The success of a backdoor attack can be generally evaluated by clean data accuracy (CDA) and attack success rate (ASR) \citep{bagdasaryan2020backdoor}. For a successful backdoor model $f^b(\theta)$, the model has a high ASR, while the CDA should be similar to the clean model $f(\theta)$.

\begin{itemize}
\item \textbf{Clean Data Accuracy (CDA)}: The CDA is the
proportion of clean test samples containing no trigger $x^{cl}_{test}$ that is
correctly predicted to their ground-truth classes $y^{cl}_{test}$.
\item \textbf{Attack Success Rate (ASR)}: The ASR is the
proportion of test samples with stamped the trigger $x^{b}_{test}$ that is predicted to the attacker targeted classes $y^{b}_{test}$.
\end{itemize}

We follow the training experimental setting of \citet{wang2019neural}:
\begin{itemize}
\item We use two forms of backdoor attacks, BadNet Attack and Complex Trigger Attack (Backdoor (\Rmnum{1}) and Backdoor (\Rmnum{1})).
\item We use a trigger $k$ size of 4 $\times$ 4 pixels.
\item We use the proportion of backdoor data $\alpha$ of 0.05.
\end{itemize}

\section{Experiments}
\subsection{Experiment for existence of key neurons}
\label{appendix:key_neurons}

\begin{figure}[h]
    \centering
    \includegraphics[width=1\textwidth]{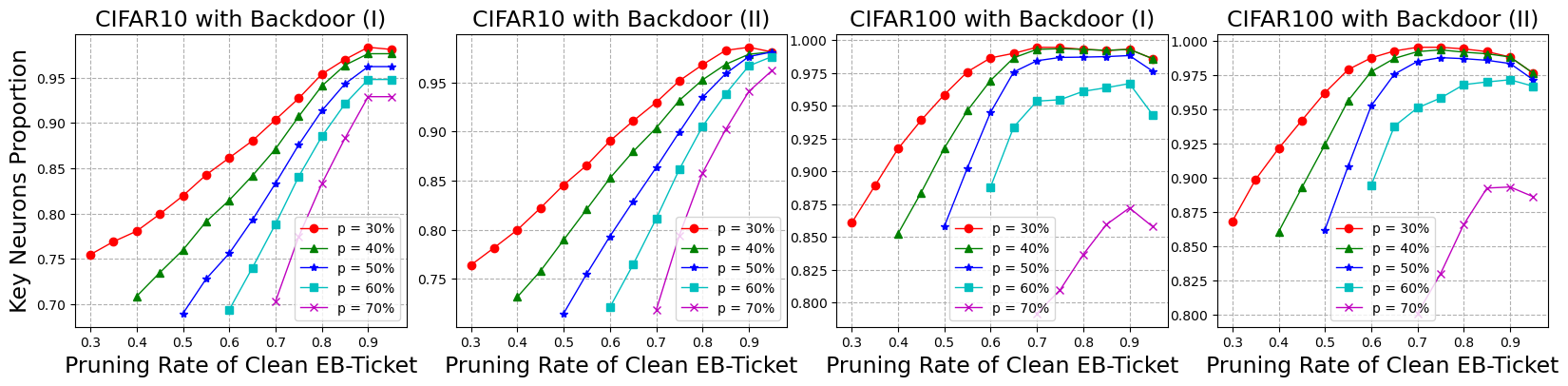}
    \caption{The neuron prune relation between the normal VGG16 model $f(\theta)$ and backdoor model $f^b(\theta)$.}
    \label{neurons_vgg}
\end{figure}

\begin{figure}[H]
    \centering
    \includegraphics[width=1\textwidth]{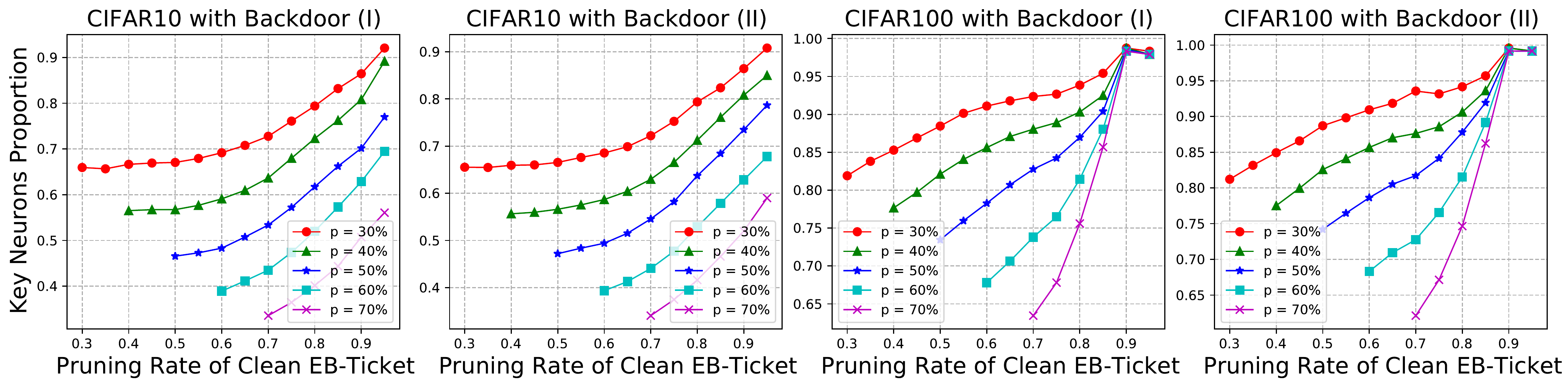}
    \caption{The neuron prune relation between the normal RESNET18 model $f(\theta)$ and backdoor model $f^b(\theta)$.}
    \label{neurons_resnet}
\end{figure}

We gradually increase the pruning rate of the normal model $f(\theta)$ to retain more important neurons and judge whether the backdoor tickets contain these neurons. 
As shown in \figurename~\ref{neurons_vgg} and \figurename~\ref{neurons_resnet}, each line represents one backdoor model after drawing EB tickets under the given pruning rate \emph{p} while the points on this line represent the proportion that the different pruning rate's key neurons in $f(\theta)$ also exist in $f^b(\theta)$.
The line shows an upward trend compared with the more important neurons, which further verifies our suppose on the two model VGG16 and ResNet18 that most neurons and connections that play a leading role in the network account for a small proportion and are preserved in both normal subnetworks and backdoor subnetworks. 

Experiments show that the more important nodes are indeed retained in the backdoor tickets with a greater probability, thus affected tickets can still achieve high accuracy.

\subsection{Experiment for intensity of attacks}


The hyperparameter $\alpha$, the proportion of backdoor data in the dataset $\mathcal{D}^{b}$, plays an important role in the attack effect. Obviously, a larger $\alpha$ makes the attack more successful. For different backdoor attacks, the attacker needs to set different values for model training to make the attacks effective \citep{chen2017targeted}. 

\begin{figure}[H]
  \centering
  \includegraphics[width=1\textwidth]{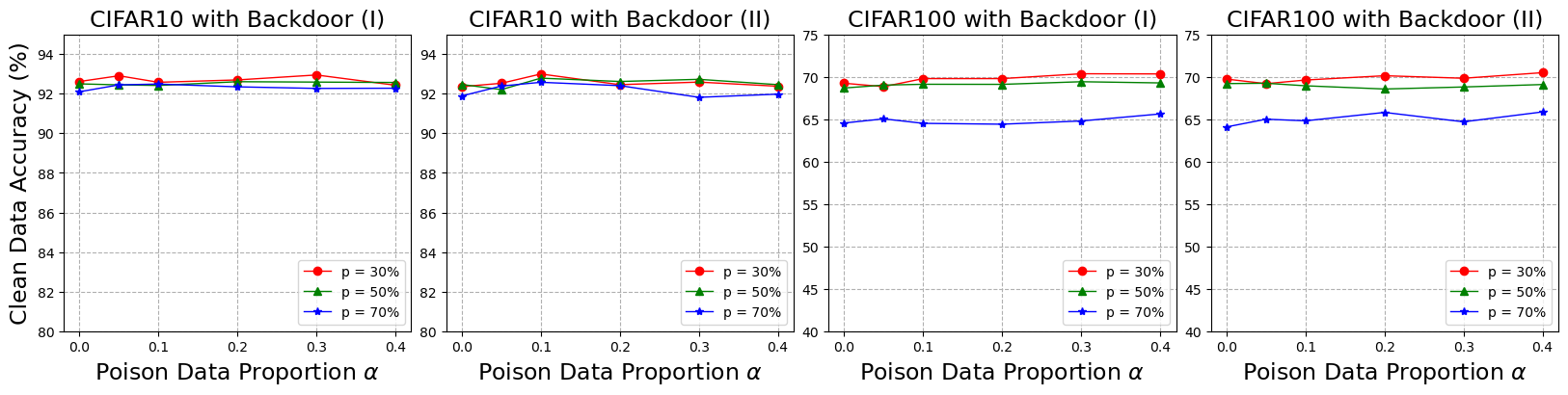}
  \caption{The retraining accuracy of subnetworks drawn under different intensities of backdoor attacks (during searching procedure), where benign datasets are used to evaluate the performance of EB tickets in retrain process.}
  \label{attack_intensity}
\end{figure}

\figurename~\ref{attack_intensity} shows that the winning tickets drawn under different intensities (different backdoor data proportion) of backdoor attacks can achieve comparable accuracy to the normal subnetworks, which shows that backdoor attacks can not obviously affect the model's performance of normal retraining. The reason why affected EB tickets can still achieve high accuracy is thought-provoking. Interestingly, this makes the backdoor attack on EB Ticket more covert because we hardly distinguish the attack by CDA's change.

